\documentclass[10pt,twocolumn,letterpaper]{article}

\usepackage{cvpr}
\usepackage{times}
\usepackage{epsfig}
\usepackage{graphicx}
\usepackage{amsmath}
\usepackage{amssymb}
\usepackage{array}

\usepackage{algorithm}
\usepackage[noend]{algorithmic}

\def\methodname{OICR}

\newcolumntype{L}[1]{>{\raggedright\let\newline\\\arraybackslash\hspace{0pt}}m{#1}}
\newcolumntype{R}[1]{>{\raggedleft\let\newline\\\arraybackslash\hspace{0pt}}m{#1}}
\newcolumntype{C}[1]{>{\centering\let\newline\\\arraybackslash\hspace{0pt}}m{#1}}

\newcolumntype{x}{>\small c}

\usepackage[pagebackref=true,breaklinks=true,letterpaper=true,colorlinks,bookmarks=false]{hyperref}

\cvprfinalcopy 

\def\cvprPaperID{1036} 

\ifcvprfinal\fi
\begin{document}

\title{Multiple Instance Detection Network with Online Instance Classifier Refinement}

\author{Peng Tang
\quad
Xinggang Wang
\quad
Xiang Bai
\quad
Wenyu Liu\thanks{Corresponding author.}\\
School of EIC, Huazhong University of Science and Technology\\
{\tt\small \{pengtang,xgwang,xbai,liuwy\}@hust.edu.cn}
}

\maketitle

\begin{abstract}
   Of late, weakly supervised object detection is with great importance in object recognition.
   Based on deep learning, weakly supervised detectors have achieved many promising results.
   However, compared with fully supervised detection, it is more challenging to train deep network based detectors in a weakly supervised manner.
   Here we formulate weakly supervised detection as a Multiple Instance Learning (MIL) problem, where instance classifiers (object detectors) are put into the network as hidden nodes.
   We propose a novel online instance classifier refinement algorithm to integrate MIL and the instance classifier refinement procedure into a single deep network, and train the network end-to-end with only image-level supervision, \ie, without object location information.
   More precisely, instance labels inferred from weak supervision are propagated to their spatially overlapped instances to refine instance classifier online.
   The iterative instance classifier refinement procedure is implemented using multiple streams in deep network, where each stream supervises its latter stream.
   Weakly supervised object detection experiments are carried out on the challenging PASCAL VOC 2007 and 2012 benchmarks.
   We obtain $47\%$ mAP on VOC 2007 that significantly outperforms the previous state-of-the-art.
\end{abstract}


\section{Introduction}
\label{sec:intro}

With the development of Convolutional Neural Network (CNN) \cite{Ref:Krizhevsky2012,Ref:Lecun1998}, great improvements have been achieved on object detection \cite{Ref:Girshick2015,Ref:Girshick2016,Ref:Liu2016,Ref:Redmon2016,Ref:Ren2015}, due to the availability of large scale datasets with accurate boundingbox-level annotations \cite{Ref:Deng2009,Ref:Everingham2010,Ref:Lin2014}.
However, collecting such accurate annotations can be very labor-intensive and time-consuming, whereas achieving only image-level annotations (\ie, image tags) is much easier, as these annotations are often available at the Internet (\eg, image search queries \cite{Ref:Li2013}).
In this paper, we aim at the Weakly Supervised Object Detection (WSOD) problem, \ie, only image tags are available during training to indicate whether an object exists in an image.

Most of previous methods follow the Multiple Instance Learning (MIL) pipeline for WSOD \cite{Ref:Bilen2015,Ref:Bilen2016,Ref:Cibis2017,Ref:Kantorov2016,Ref:Song2014,Ref:Wang2014,Ref:Wang2015}.
They treat images as bags and image regions generated by object proposal methods \cite{Ref:Uijlings2013,Ref:Zitnick2014} as instances to train instance classifiers (object detectors) under the MIL constraints \cite{Ref:Dietterich1997}. 
Meanwhile, recent efforts tend to combine MIL and CNN by either using CNN as an off-the-shelf feature extractor \cite{Ref:Bilen2015,Ref:Cibis2017,Ref:Song2014,Ref:Wang2014,Ref:Wang2015} or training an end-to-end MIL network \cite{Ref:Bilen2016,Ref:Kantorov2016}.
Here we are also along the MIL line for WSOD, and train an end-to-end network.

\begin{figure}[t]
\begin{center}
   \includegraphics[width=0.95\linewidth]{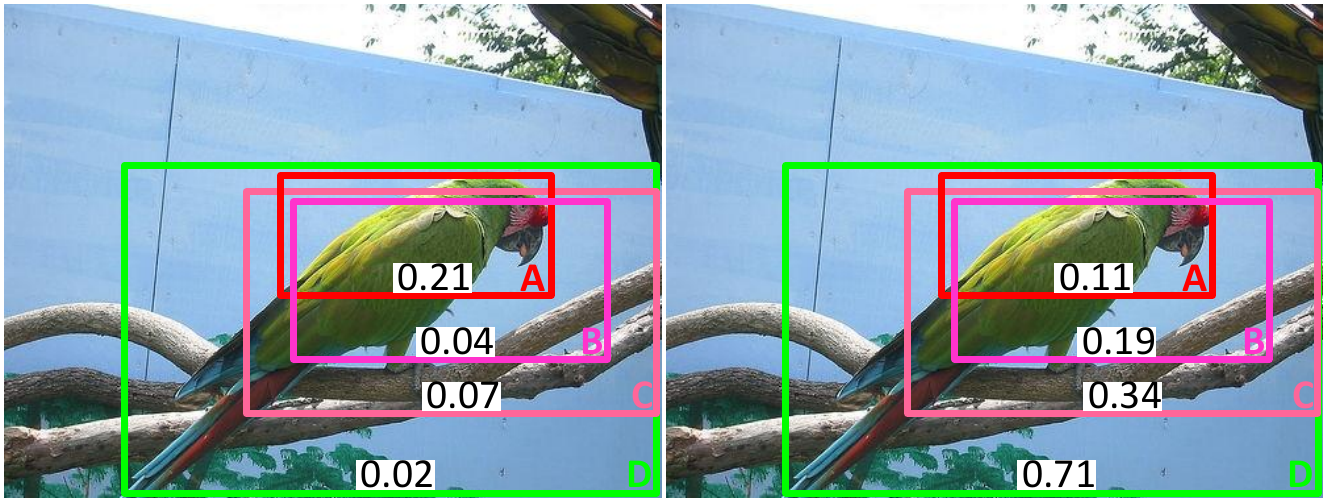}
\end{center}
   \caption{Detection results without/with classifier refinement (left/right).
   Detection scores are plotted in the bottom of the sampled proposals A, B, C, and D.
   In the left, the top ranking proposal A does not correctly localize the object.
   After instance classifier refinement, in the right, the correct proposal D is detected and more discriminative performance of instance classifier is shown.}
\label{fig:visualize0}
\end{figure}

\begin{figure*}[t]
\begin{center}
   \includegraphics[width=0.9\linewidth]{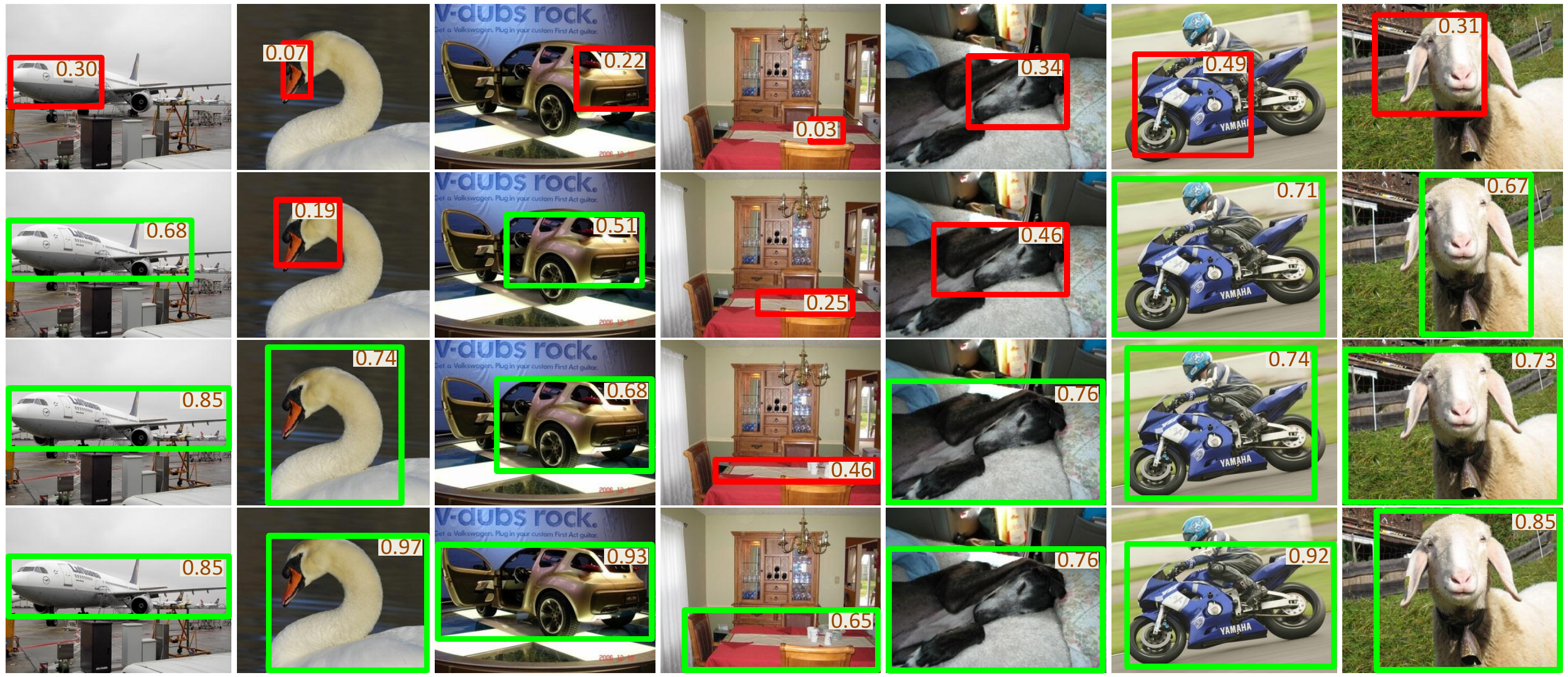}
\end{center}
   \caption{Detection results from different stages of classifier refinement.
   Each row represents one stage.
   Green/red rectangles indicate detected boxes having high/low overlap with ground truths, and digits in the top right corner of rectangles indicate the IoU.
   Through multi-stage refinement, the detector can cover the whole object instead of parts gradually.}
\label{fig:visualize}
\end{figure*}

Though many promising results have been achieved in WSOD, they are still far from comparable to fully supervised ones \cite{Ref:Girshick2015,Ref:Girshick2016,Ref:Ren2015}.
Weakly supervised object detection only requires supervision at image category level. 
Bilen and Vedaldi \cite{Ref:Bilen2016} presents an end-to-end deep network for WSOD, in which final image classification score is the weighted sum of proposal scores, that is, each proposal contributes a percentage to the final image classification.
The deep network can correctly classify image even only ``see'' a part of object, and as a result, the top ranking proposal may fail to meet the standard object detection requirement (IoU$>$0.5 between ground truths and predicted boxes).
As shown in Fig.~\ref{fig:visualize0}~(left), the top-ranking proposal A is too small.
Meanwhile, proposals B, C, and D have similar detection scores.
This shows that the WSOD network is not discriminative enough to correctly localize object.
This is a core problem of end-to-end deep network based WSOD.
To address this problem, we put forward two improvements in this paper:
1) Instead of estimating instance weights through weighted sum pooling, we propose to add some blocks in the network for learning more discriminative instance classifiers by explicitly assigning binary instance labels;
2) We propose to refine instance classifier online using spatial relation.

Our motivation is that, though some detectors only capture objects partially, proposals having high spatial overlaps with detected parts may cover the whole object, or at least contain larger portion of the object.
In \cite{Ref:Bilen2016}, Bilen and Vedaldi propose a spatial regulariser via forcing features of highest scoring region and its adjacent regions to be the same, which significantly improves WSOD performance.
Nevertheless, forcing spatially overlapped proposals to have the same features seems too rigorous.
Rather than taking the rigorous constraint, we think the features of spatially overlapped proposals are in the same manifold.
Then these overlapped proposals could share similar label information.
As shown in Fig.~\ref{fig:visualize0}~(right), we except the label information of A can propagate to B and C which has large overlap with A, and then the label information of B and C can propagate to D to correctly localize object. 
To implement this idea, we design some instance classifiers in the network of \cite{Ref:Bilen2016}. 
The labels of instance could be refined by their spatially overlapped instances.
We name this new network structure Multiple Instance Detection Network (MIDN) with instance classifier.

In practice, there are two important issues.
1) How to initialize instance labels, since there is no instance-level supervision in this task.
2) How to train the network with instance classifier efficiently.
A natural way for classifier refinement is the alternative strategy, that is, alternatively relabelling instance and training instance classifier, while this procedure is very time-consuming, especially considering training deep networks with a huge number of Stochastic Gradient Descent (SGD) iterations.
To overcome these difficulties, we propose a novel Online Instance Classifier Refinement (OICR) algorithm to train the network online.

Our method has multiple output streams for different stages: the first is the MIDN to train a basic instance classifier and others refine the classifier.
To refine instance classifier online, after the forward process of SGD, we can obtain a set of proposal scores.
According to these scores, for each stage, we can label the top-scoring proposal along with its spatially overlapped proposals to the image label.
Then these proposal labels can be used as the supervision to train instance classifier in the next stage.
Though the top-scoring proposal may only contain a part of an object, its adjacent proposals will cover larger portion of the object.
Thus the instance classifier can be refined.
After implementing the refinement procedure multiple times, the detector can discover the whole object instead of parts gradually, as shown in Fig.~\ref{fig:visualize}.
But in the beginning of training, all classifiers are almost non-trained, which will result in very noisy supervision of refined classifiers, and then the training will deviate from correct solutions a lot.
To solve this problem, we design a weighted loss further by assigning different weights to different proposals in different training iterations.
Using this strategy, all classifier refinement procedures can thus be integrated into a single network and trained end-to-end.
It can improve the performance benefiting from the classifier refinement procedure.
Meanwhile, the multi-stage strategy and online refinement algorithm is very computational efficient in both training and testing.
Moreover, performance can be improved by sharing representations among different training stages.

\begin{figure*}[!t]
\begin{center}
   \includegraphics[width=0.89\linewidth]{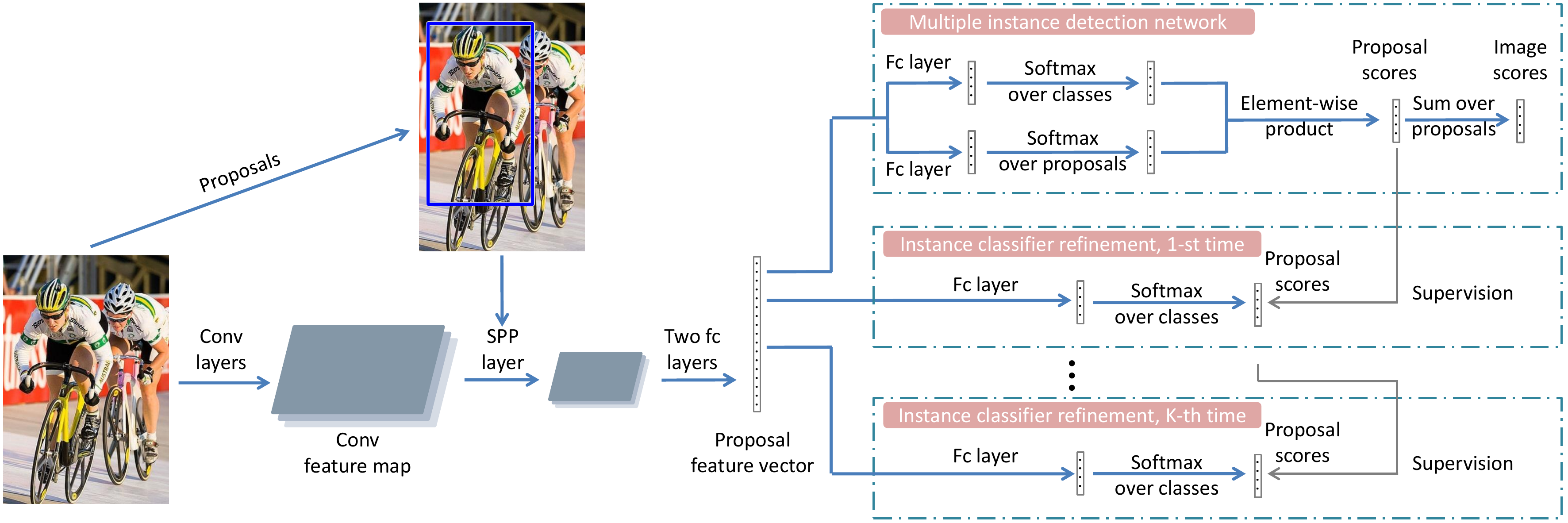}
\end{center}
   \caption{The architecture of MIDN with OICR.
   Proposal/instance feature is generated by the spatial pyramid pooling layer on the convolutional feature map of image and two fully connected layers.
   These proposal feature vectors are branched into many streams for different stages: the first one for the basic multiple instance detection network and others for instance classifier refinement.
   Supervision for classifier refinement is decided by outputs from their preceding stages.
   All these stages share the same proposal representations.}
\label{fig:architecture}
\end{figure*}

We elaborately conduct many experiments on the challenging PASCAL VOC dataset to confirm the effectiveness of our method.
Our method achieves $47.0\%$ mAP and $64.3\%$ CorLoc on VOC 2007 that outperforms previous best performed methods by a large margin.

In summary, the main contributions of our work are listed as follows.
\begin{itemize}
   \item We propose a framework for weakly supervised learning that combines MIDN with multi-stage instance classifiers.
   With only supervision of the outputs from its preceding stage, the discriminatory power of the instance classifier can be enhanced iteratively.
   \item We further design a novel OICR algorithm that integrates the basic detection network and the multi-stage instance-level classifier into a single network.
   The proposed network is end-to-end trainable.
   Compared with the alternatively training strategy, we demonstrate that our method can not only reduce the training time, but also boost the performance.
   \item Our method achieves significantly better results over previous state-of-the-art methods on the challenging PASCAL VOC 2007 and 2012 benchmarks for weakly supervised object detection.
\end{itemize}

\section{Related work}
\label{sec:related_work}

MIL is a classical weakly supervised learning problem and was first proposed in \cite{Ref:Dietterich1997} for drug activity prediction.
After that, many solutions have been proposed for MIL \cite{Ref:Andrews2002,Ref:Wang2015,Ref:Zhang2001}.
In MIL, a set of bags are given, and each bag is associated with a collection of instances.
MIL has two constraints:
1) If a bag is positive, at least one instance in the bag is positive;
2) If a bag is negative, all instances in the bag are negative.
It is natural to treat WSOD as a MIL problem.
Then the problem turns into finding an instance classifier only given bag labels.
Our method also follows the MIL line, and the classifier refinement is inspired by the classifier updating procedure in mi-SVM \cite{Ref:Andrews2002} to some extent.
The differences are that, in mi-SVM, it uses an alternative strategy to relabel instances and retrain a classifier, while we adopt an online refinement algorithm; the mi-SVM relabel instances according to the instance score predicted by the classifier, while we select instances according to the spatial relation.

Most of the existing methods solve the WSOD problem based on MIL \cite{Ref:Bilen2015,Ref:Bilen2016,Ref:Cibis2017,Ref:Kantorov2016,Ref:Oquab2015,Ref:Song2014,Ref:Wang2015}.
For example, Wang \etal \cite{Ref:Wang2015} relaxed the MIL restraints into a differentiable loss function and optimized it by SGD to speed up training and improve results.
Cibis \etal \cite{Ref:Cibis2017} trained a multi-fold MIL detector by alternatively relabelling instances and retraining classifier.
Recently, some researchers combined CNN and MIL to train an end-to-end network for WSOD \cite{Ref:Bilen2016,Ref:Kantorov2016,Ref:Oquab2015}.
Oquab \etal \cite{Ref:Oquab2015} trained a CNN network using the max-pooing MIL strategy to localize objects.
But their methods can only coarsely localize objects regardless of their sizes and aspect ratios, our method can detect objects more accurately.
Bilen and Vedaldi \cite{Ref:Bilen2016} proposed a Weakly Supervised Deep Detection Network (WSDDN), which presents a novel weighted MIL pooling strategy and combines with the proposal objectness and spatial regulariser for better performance.
Based on the WSDDN, Kantorov \etal \cite{Ref:Kantorov2016} used a contrastive model to consider the context information for improvement.
We also choose the WSDDN as our basic network, but we combine it with multi-stage classifier refinement, and propose a novel OICR algorithm to train our network effectively and efficiently, which can boost the performance significantly.
Different from the spatial regulariser in WSDDN \cite{Ref:Bilen2016} that forces features of highest scoring proposal and its spatially overlapped proposals to be the same, our OICR assumes features of spatially overlapped proposals are in the same manifold, which is more reasonable.
Experiments on Section~\ref{sec:exp} demonstrate that our strategy can obtain more superior results.

The proposal labelling procedure is also related to the semi-supervised label propagation method \cite{Ref:Bai2017,Ref:Zhu2002}.
But in label propagation, it labels data according to the similarity among labelled and unlabelled data, while we use spatial overlap as the metric; and there are no available labelled instances for propagation, which is quite different from semi-supervised methods.
Meanwhile, the sharing representation strategy in our network is similar to multi-task learning \cite{Ref:Caruana1997}.
Unlike the multi-task learning that each output stream has their own relatively independent external supervision, in our method, supervision of latter streams only depends on the outputs from their preceding streams.

\section{Method}
\label{sec:method}

The overall architecture of our method is shown in Fig.~\ref{fig:architecture}.
Given an image, we first generate about $2,000$ object proposals by Selective Search \cite{Ref:Uijlings2013}.
The image and these proposals are fed into some convolutional (conv) layers with Spatial Pyramid Pooling (SPP) layer \cite{Ref:He2015} 
to produce a fixed-size conv feature map per-proposal,
and then they are fed into two fully connected (fc) layers to generate a collection of proposal feature vectors.
These features are branched into different streams, \ie, different stages: the first one is the MIDN to train a basic instance classifier and others refine classifier.
Specially, supervision for classifier refinement is decided by outputs from their preceding stages, \eg, supervision of the first refined classifier depends on the output from the basic classifier, and supervision of $k^{\textup{th}}$ refined classifier depends on outputs from $\{k-1\}^{\textup{th}}$ refined classifier.

In this section, we will introduce the chosen basic MIDN, and explain our \methodname\ algorithm in detail.

\subsection{Multiple instance detection network}
\label{sec:midn}

It is necessary to achieve instance-level supervision to train refined classifier, yet such supervision is unavailable.
As we have stated before, the top-scoring proposal by instance classifiers and its adjacent proposals can be labelled to its image label as supervision.
So we first introduce our MIDN to generate the basic instance classifier.
There are many possible choices \cite{Ref:Bilen2016,Ref:Cibis2017,Ref:Kantorov2016,Ref:Wang2015} to achieve this.
Here we choose the method by Bilen and Vedaldi \cite{Ref:Bilen2016} which proposes a weighted pooling strategy to obtain the instance classifier, for its effectiveness and implementation convenience.
Notice that our network is independent of special MIL methods, so any method that can be trained end-to-end could be embedded into our network.

As shown in the ``Multiple instance detection network'' block of Fig.~\ref{fig:architecture}, proposal features are branched into two streams to produce two matrices $\mathbf{x}^{c}, \mathbf{x}^{d} \in \mathbb{R}^{C \times |R|}$ of image by two fc layers, where $C$ denotes the number of image classes and $|R|$ denotes the number of proposals.
Then the two matrices are passing through two softmax layer along different directions:
$[\sigma(\mathbf{x}^{c})]_{ij} = \frac{e^{x_{ij}^{c}}}{\sum_{k=1}^{C} e^{x_{kj}^{c}}}$ and $[\sigma(\mathbf{x}^{d})]_{ij} = \frac{e^{x_{ij}^{d}}}{\sum_{k=1}^{|R|} e^{x_{ik}^{d}}}$.
The proposal scores are generated by element-wise product $\mathbf{x}^{R} = \sigma(\mathbf{x}^{c}) \odot \sigma(\mathbf{x}^{d})$.
At last, image score of $c^{\textup{th}}$ class $\phi_{c}$ can be obtained by the sum over all proposals: $\phi_{c} = \sum_{r=1}^{|R|} x_{cr}^{R}$.

The interpretation of the two streams framework is as follows.
The $[\sigma(\mathbf{x}^{c})]_{ij}$ is the probability of proposal $j$ belonging to class $i$.
The $[\sigma(\mathbf{x}^{d})]_{ij}$ is the normalized weight that indicates the contribution of proposal $j$ to image being classified to class $i$.
So $\phi_{c}$ is achieved by weighted sum pooling and falls in the range of $(0, 1)$.
Given image label $\mathbf{Y} = [y_{1}, y_{2}, ..., y_{C}]^{T} \in \mathbb{R}^{C\times1}$, where $y_{c} = 1$ or $0$ indicates the image with or without object $c$.
We can train the basic instance classifier by standard multi-class cross entropy loss, as shown in Eq.~(\ref{equ:loss_base}), then the instance classifier can be obtained according to the proposal score $\mathbf{x}^{R}$.
More details can be found in \cite{Ref:Bilen2016}.

\begin{equation}
\label{equ:loss_base}
   \textup{L}_\textup{b} = -\mathop{\sum} \limits_{c=1}^{C} \{y_{c} \log \phi_{c} + (1 - y_{c}) \log (1 - \phi_{c})\}.
\end{equation}

\subsection{Online instance classifier refinement}
\label{sec:oicr}

In the last subsection, we have obtained the basic instance classifier.
Here we will expound how to refine instance classifiers online.
A natural way to refine classifier is an alternative strategy, that is, fixing the classifier and labelling proposals, fixing proposal labels and training the classifier.
But it has some limitations:
1) It is very time-consuming as it requires training the classifier multiple times;
2) Training different classifiers in different refinement steps separately may harm the performance because it hinders the process to benefit from the shared representations.
Hence, we integrate the basic MIDN and different classifier refinement stages into a single network and train it end-to-end.

The difficulty is how to obtain instance labels for refinement when there are no available labelled instances.
To deal with this problem, we propose an online labelling and refinement strategy.
Different from the basic instance classifier, the output score vector $\mathbf{x}^{Rk}_{j}$ of proposal $j$ for refined classifier is a $\{C+1\}$-dimensional vector, \ie, $\mathbf{x}^{Rk}_{j} \in \mathbb{R}^{(C+1)\times1}, k \in \{1, 2, ..., K\}$, where the $k$ is for $k^{\textup{th}}$ time refinement, $K$ is the total refinement times, and the $\{C+1\}^{\textup{th}}$ dimension is for background (here we represent the proposal score vector from the basic classifier as $\mathbf{x}^{R0}_{j} \in \mathbb{R}^{C\times1}$).
The $\mathbf{x}^{Rk}_{j}, k>0$ is obtained by passing the proposal feature vector through a single fc layer and a softmax over classes layer, as shown in the ``Instance classifier refinement'' block of Fig.~\ref{fig:architecture}.

Suppose the label vector for proposal $j$ is $\mathbf{Y}^{k}_{j} = [y^{k}_{1j}, y^{k}_{2j}, ..., y^{k}_{(C+1)j}]^{T} \in \mathbb{R}^{(C+1) \times 1}$.
In each training iteration, after the forward process of SGD, we can get a set of proposal scores $\mathbf{x}^{R(k-1)}$.
Then we can obtain the supervision of refinement time $k$ according to $\mathbf{x}^{R(k-1)}$.
There are many possible methods to obtain instance labels using $\mathbf{x}^{R(k-1)}$, \eg, labeling an instance as positive if its score exceeds a threshold, otherwise as negative, as the mi-SVM \cite{Ref:Andrews2002}.
But in our case, the score for each instance is changed during each training iteration, and for different classes, using the same threshold may not be suitable, thus it is hard to settle a threshold.
Here we choose a different strategy, inspired by the fact that highly spatially overlapped instances should have the same label.
Suppose an image has class label $c$, we first select proposal $j^{k-1}_{c}$ with highest score for $\{k-1\}^{\textup{th}}$ time as in Eq.~(\ref{equ:j}), and label it to class $c$, \ie, $y^{k}_{cj^{k-1}_{c}} = 1$ and $y^{k}_{c^{\prime}j^{k-1}_{c}} = 0, c^{\prime} \ne c$.
As different proposals always have overlaps, and proposals with high overlap should belong to the same class, we can label proposal $j^{k-1}_{c}$ and its adjacent proposals to class $c$ for $k^{\textup{th}}$ refinement, \ie, if proposal $j$ have a high overlap with proposal $j^{k-1}_{c}$, we label proposal $j$ to class $c$ ($y^{k}_{cj} = 1$), otherwise we label proposal $j$ as background ($y^{k}_{(C+1)j} = 1$).
Here we label proposal $j$ to class $c$ if the IoU between proposal $j$ and $j^{k-1}_{c}$ greater than a threshold $I_{t}$ which is determined by experiments.
Meanwhile, if there is no object $c$ in the image, we set all $y^{k}_{cj} = 0$.
Using this supervision, we can train the refined classifier based on the loss function in Eq.~(\ref{equ:loss_refine}).
Through multiple times of refinement, our detector can detect larger parts of objects gradually.

\begin{equation}
\label{equ:j}
   j^{k-1}_{c} = \textup{arg}\max_{r} x^{R(k-1)}_{cr}.
\end{equation}

\begin{equation}
\label{equ:loss_refine}
   \textup{L}^{k}_\textup{r} = -\frac{1}{|R|}\mathop{\sum} \limits_{r=1}^{|R|} \sum_{c=1}^{C+1} y^{k}_{cr} \log x^{Rk}_{cr}.
\end{equation}

\begin{algorithm}[t]
\caption{Online instance classifier refinement}
\label{alg:oicr}
\begin{algorithmic}[1]
   \REQUIRE Image $X$ and its proposals; image label vector $\mathbf{Y} = [y_{1}, ..., y_{C}]$; refinement times $K$.
   \ENSURE Loss weights $w^{k}_{r}$; proposal label vectors $\mathbf{Y}^{k}_{r} = [y^{k}_{1r}, ..., y^{k}_{(C+1)r}]^{T}$. Where $r \in \{1, ..., |R|\}$ and $k \in \{1, ..., K\}$.
   \STATE Feed $X$ and its proposals into the network to produce proposal score matrices  $\mathbf{x}^{Rk}$, $k \in \{0, ..., K-1\}$.
   \FOR{$k=0$ \TO $K-1$}
   \STATE Set all elements in $\mathbf{I} = [I_{1}, ..., I_{|R|}]^{T}$ to $-\inf$.
   \STATE Set all $y^{k+1}_{cr}=0, c \in \{1, ..., C\}$ and $y^{k+1}_{(C+1)r}=1$.
      \FOR{$c=1$ \TO $C$}
         \IF{$y_{c} = 1$}
            \STATE  Choose the top-scoring proposal $j^{k}_{c}$ by Eq.~(\ref{equ:j}).
            \FOR{$r=1$ \TO $|R|$}
               \STATE Compute IoU $I^{\prime}_{r}$ between proposal $r$ and $j^{k}_{c}$.
               \IF{$I^{\prime}_{r} > I_{r}$}
                  \STATE Set $I_{r} = I^{\prime}_{r}$ and $w^{k+1}_{r} = x^{Rk}_{cj^{k}_{c}}$.
                  \IF{$I_{r} > I_{t}$}
                     \STATE Set $y^{k+1}_{c^{\prime}r}=0, c^{\prime} \ne c$ and $y^{k+1}_{cr} = 1$.
                  \ENDIF
               \ENDIF
            \ENDFOR
         \ENDIF
      \ENDFOR
   \ENDFOR
\end{algorithmic}
\end{algorithm}

Actually the acquired supervision for refining classifier is very noisy, especially in the beginning of training, which will result in unstable solutions.
To solve this problem, we change the loss in Eq.~(\ref{equ:loss_refine}) to a weighted version, as in Eq.~(\ref{equ:loss_refine_weighted}).
\begin{equation}
\label{equ:loss_refine_weighted}
   \textup{L}^{k}_\textup{r} = -\frac{1}{|R|}\mathop{\sum} \limits_{r=1}^{|R|} \sum_{c=1}^{C+1} w^{k}_{r} y^{k}_{cr} \log x^{Rk}_{cr},
\end{equation}
where $w^{k}_{r}$ is the loss weight and can be acquired by the $11^{\textup{th}}$ line of Algorithm~\ref{alg:oicr}.
The explanation of such choice is as follows.
In the beginning of training, the $w^{k}_{r}$ is small, hence, the loss is also small.
As a consequence, the performance of the network will not decrease a lot though good positive instances cannot be found.
Meanwhile, during the training procedure, the network can achieve positive instances with high scores easily for easy bags, and these positive instances are always with high scores, \ie, $w^{k}_{r}$ is large.
On the contrary, it is difficult to get positive instances for difficult bags, as a result, these positive instances are always very noisy.
Nevertheless, the refined classifier will not deviate from the correct solution a lot, because the scores of these noisy positive instances are relatively low, \ie, $w^{k}_{r}$ is small.

To make the \methodname\ algorithm more clear, we summarize the process to obtain supervision in Algorithm~\ref{alg:oicr}, where $I_{r}$ indicates the maximum IoU between proposal $r$ and the top-scoring proposal.
After obtaining supervision and loss for training refined classifiers, we can get the loss of our overall network by combining Eq.~(\ref{equ:loss_base}) and Eq.~(\ref{equ:loss_refine_weighted}), as Eq.~(\ref{equ:loss_all}).
Through optimizing this loss function, we can integrate the basic network and different classifier refinement stages into a single network, and share representations among different stages.

\begin{equation}
\label{equ:loss_all}
   \textup{L} = \textup{L}_{\textup{b}} + \mathop{\sum} \limits_{k=1}^{K} \textup{L}^{k}_{r}.
\end{equation}

\section{Experiments}
\label{sec:exp}

\subsection{Experimental setup}
\label{sec:exp_setup}

In this section we will perform thorough experiments to analyse our \methodname\ and its components for weakly supervised object detection.

\paragraph{Datasets and evaluation measures}
We evaluate our method on the challenging PASCAL VOC 2007 and 2012 datasets \cite{Ref:Everingham2010} which have $9,962$ and $22,531$ images respectively for $20$ object classes.
These two datasets are divided into train, val, and test sets.
Here we choose the trainval set ($5,011$ images for 2007 and $11,540$ for 2012) to train our network.
As we focus on weakly supervised detection, only image-level labels are utilized during training.
For testing, there are two metrics for evaluation: mAP and CorLoc.
Average Precision (AP) and the mean of AP (mAP) is the evaluation metric to test our model on the testing set, which follows the standard PASCAL VOC protocol \cite{Ref:Everingham2010}.
Correct localization (CorLoc) is to test our model on the training set measuring the localization accuracy \cite{Ref:Deselaers2012}.
All these two metrics are based on the PASCAL criteria, \ie, IoU$>$0.5 between ground truths and predicted boxes.

\begin{figure}[t]
\begin{center}
   \includegraphics[width=0.85\linewidth]{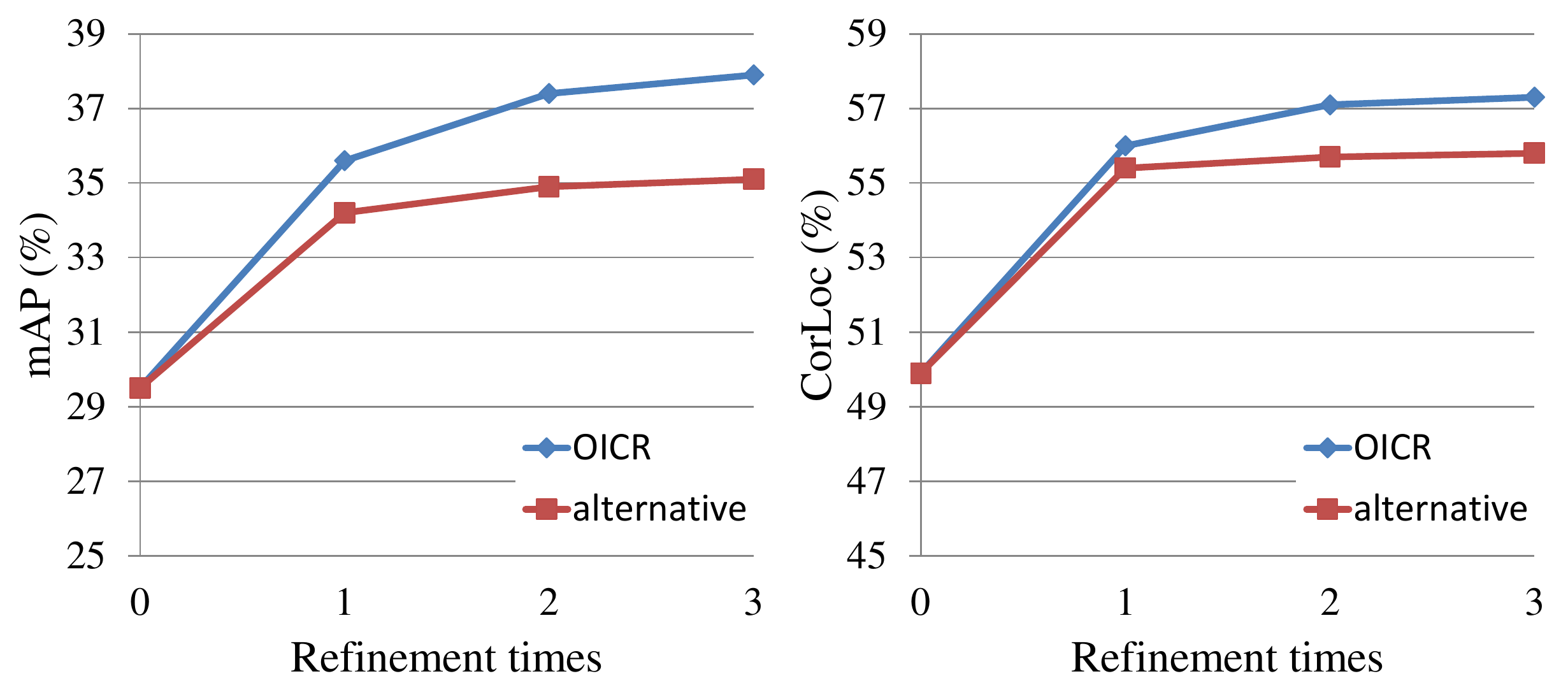}
\end{center}
   \caption{Results on VOC 2007 for different refinement times and different training strategies, where ``\methodname'' indicates our \methodname\ training strategy, ``alternative'' indicates the alternative strategy.}
\label{fig:ablation_refinement_time}
\end{figure}

\begin{table}[t]
\begin{center}
\footnotesize
\begin{tabular}{|l|c|c|}
   \hline
   Loss & mAP ($\%$) & CorLoc ($\%$) \\
   \hline\hline
   unweighted & 32.8 & 50.6 \\
   weighted & \bf{37.9} & \bf{57.3} \\
   \hline
\end{tabular}
\end{center}
\caption{Results on VOC 2007 for different losses.}
\label{table:weight}
\end{table}

\paragraph{Implementation details}
Our method is built on two pre-trained ImageNet \cite{Ref:Deng2009} networks: VGG$\_$M \cite{Ref:Chatfield2014} and VGG16 \cite{Ref:Simonyan2015}, each of which has some conv layers with max-pooling layer and three fc layers.
We replace the last max-pooling layer of the two models by SPP layer, and the last fc layer and softmax loss layer by the layers described in Section~\ref{sec:method}.
To increase the feature map size from the last conv layer, we replace the penultimate max-pooling layer and its subsequent conv layers by the dilated conv layers \cite{Ref:Yu2016}.
The new added layers are initialized using Gaussian distributions with $0$-mean and standard deviations $0.01$.
Biases are initialized to $0$.
During training, the mini-batch size for SGD is set to $2$, and the learning rate is set to $0.001$ for the first $40$K iterations and then decrease to $0.0001$ in the following $30$K iterations.
The momentum and weight decay are set to $0.9$ and $0.0005$ respectively.

As we have stated in Section~\ref{sec:method}, Selective Search (SS) \cite{Ref:Uijlings2013} is adopted to generate about $2,000$ proposals per-image.
For data augmentation, we use five image scales $\{480, 576, 688, 864, 1200\}$ (resize the shortest side to one of these scales) and cap the longest image side to less than $2000$ with horizontal flips for both training and testing.
We refine instance classifier three times, \ie, $K=3$ in Section~\ref{sec:oicr}, so there are four stages in total.
The IoU threshold $I_{t}$ in the $12^{\textup{th}}$ line of Algorithm~\ref{alg:oicr} is set to $0.5$.
During testing, the mean output of these three refined classifiers is chosen.
We also follow the \cite{Ref:Kumar2016,Ref:Li2016} to train a supervised object detector by choosing top-scoring proposals given by our method as pseudo ground truths to further improve our results.
Here we train a Fast RCNN (FRCNN) \cite{Ref:Girshick2015} detector using the VGG16 model and the same five image scales (horizontal flips only in training).
SS is also chosen for proposal generation to train the FRCNN.
Non-maxima suppression (with $30\%$ IoU threshold) is applied to compute AP and CorLoc.

Our experiments are implemented based on the Caffe \cite{Ref:Jia2014} deep learning framework.
All of our experiments are running on a NVIDIA GTX TitanX GPU.
Codes for reproducing the results are available at \url{https://github.com/ppengtang/oicr}.

\subsection{Ablation experiments}
\label{sec:abl_exp}

We first conduct some ablation experiments to illustrate the effectiveness of our training strategy, including the influence of classifier refinement, \methodname, weighted loss, and the IoU threshold $I_{t}$.
Without loss generality, we only perform experiments on VOC 2007 and use the VGG$\_$M model.

\begin{figure}[t]
\begin{center}
   \includegraphics[width=0.85\linewidth]{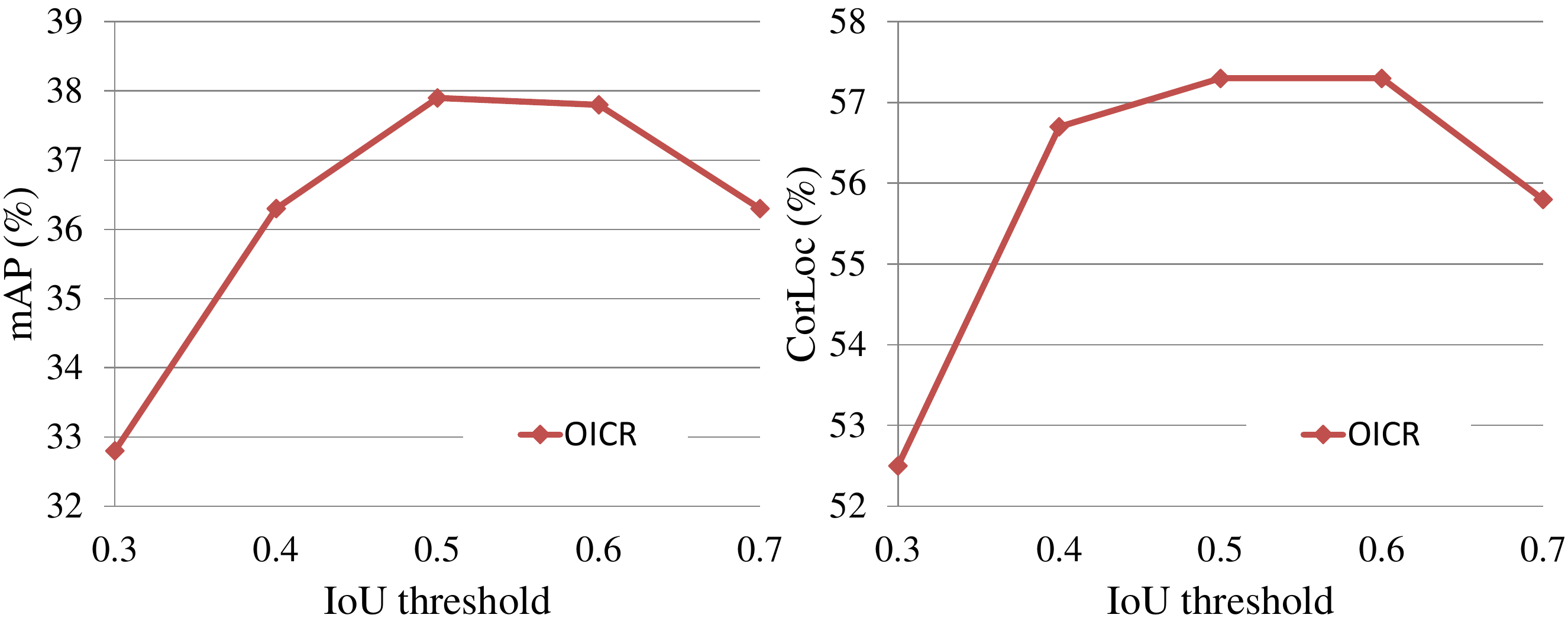}
\end{center}
   \caption{Results on VOC 2007 for different IoU threshold $I_{t}$.}
\label{fig:ablation_threshold}
\end{figure}

\begin{table*}[t]
\begin{center}
\resizebox{\linewidth}{!}{
\begin{tabular}{|@{}L{3.6cm}|*{20}{x}|x|}
   \hline
   Method & aero & bike & bird & boat & bottle & bus & car & cat & chair & cow & table & dog & horse & mbike & person & plant & sheep & sofa & train & tv & mAP \\
   \hline\hline
   WSDDN-VGG$\_$F \cite{Ref:Bilen2016} & 42.9 & 56.0 & 32.0 & 17.6 & 10.2 & 61.8 & 50.2 & 29.0 & 3.8 & 36.2 & 18.5 & 31.1 & 45.8 & 54.5 & 10.2 & 15.4 & 36.3 & 45.2 & 50.1 & 43.8 & 34.5\\
   WSDDN-VGG$\_$M \cite{Ref:Bilen2016} & 43.6 & 50.4 & 32.2 & \underline{\bf{26.0}} &  9.8 & 58.5 & 50.4 & 30.9 & 7.9 & 36.1 & 18.2 & 31.7 & 41.4 & 52.6 & 8.8 & 14.0 & 37.8 & 46.9 & 53.4 & 47.9 & 34.9\\
   WSDDN-VGG16 \cite{Ref:Bilen2016} & 39.4 & 50.1 & 31.5 & 16.3 & 12.6 & 64.5 & 42.8 & \bf{42.6} & 10.1 & 35.7 & 24.9 & 38.2 & 34.4 & 55.6 & 9.4 & 14.7 & 30.2 & 40.7 & 54.7 & 46.9 & 34.8\\
   WSDDN+context \cite{Ref:Kantorov2016} & 57.1 & 52.0 & 31.5 & 7.6 & 11.5 & 55.0 & 53.1 & 34.1 & 1.7 & 33.1 & \bf{49.2} & \underline{\bf{42.0}} & 47.3 & 56.6 & 15.3 & 12.8 & 24.8 & \bf{48.9} & 44.4 & 47.8 & 36.3\\
   \hline
   \methodname-VGG$\_$M & 53.1 & 57.1 & \bf{32.4} & 12.3 & \bf{15.8} & 58.2 & 56.7 & 39.6 & 0.9 & \bf{44.8} & 39.9 & 31.0 & \bf{54.0} & 62.4 & 4.5 & 20.6 & 39.2 & 38.1 & 48.9 & 48.6 & 37.9\\
   \methodname-VGG16 & \bf{58.0} & \bf{62.4} & 31.1 & 19.4 & 13.0 & \bf{65.1} & \bf{62.2} & 28.4 & \underline{\bf{24.8}} & 44.7 & 30.6 & 25.3 & 37.8 & \bf{65.5} & \bf{15.7} & \bf{24.1} & \bf{41.7} & 46.9 & \underline{\bf{64.3}} & \underline{\bf{62.6}} & \bf{41.2}\\
   \hline\hline
   WSDDN-Ens. \cite{Ref:Bilen2016} & 46.4 & 58.3 & 35.5 & \bf{25.9} & 14.0 & 66.7 & 53.0 & 39.2 & 8.9 & 41.8 & 26.6 & \bf{38.6} & 44.7 & 59.0 & 10.8 & 17.3 & 40.7 & 49.6 & 56.9 & 50.8 & 39.3 \\
   OM+MIL+FRCNN \cite{Ref:Li2016} & 54.5 & 47.4 & 41.3 & 20.8 & 17.7 & 51.9 & 63.5 & \underline{\bf{46.1}} & \bf{21.8} & 57.1 & 22.1 & 34.4 & 50.5 & 61.8 & \underline{\bf{16.2}} & \underline{\bf{29.9}} & 40.7 & 15.9 & 55.3 & 40.2 & 39.5\\
   \hline
   \methodname-Ens. & 58.5 & 63.0 & 35.1 & 16.9 & 17.4 & 63.2 & 60.8 & 34.4 & 8.2 & 49.7 & 41.0 & 31.3 & 51.9 & 64.8 &13.6 & 23.1 & 41.6 & 48.4 & 58.9 & 58.7 & 42.0\\
   \methodname-Ens.+FRCNN & \underline{\bf{65.5}} & \underline{\bf{67.2}} & \underline{\bf{47.2}} & 21.6 & \underline{\bf{22.1}} & \underline{\bf{68.0}} & \underline{\bf{68.5}} & 35.9 & 5.7 & \underline{\bf{63.1}} & \underline{\bf{49.5}} & 30.3 & \underline{\bf{64.7}} & \underline{\bf{66.1}} & 13.0 & 25.6 & \underline{\bf{50.0}} & \underline{\bf{57.1}} & \bf{60.2} & \bf{59.0} & \underline{\bf{47.0}}\\
\hline
\end{tabular}
}
\end{center}
\caption{Average precision (in $\%$) for different methods on VOC 2007 test set.
The upper part shows results using a single model.
The lower part shows results of combing multiple models.}
\label{table:voc_2007_map}
\end{table*}

\begin{table*}[t]
\begin{center}
\resizebox{\linewidth}{!}{
\begin{tabular}{|@{}L{3.6cm}|*{20}{x}|x|}
   \hline
   Method & aero & bike & bird & boat & bottle & bus & car & cat & chair & cow & table & dog & horse & mbike & person & plant & sheep & sofa & train & tv & mean \\
   \hline\hline
   WSDDN-VGG$\_$F \cite{Ref:Bilen2016} & 68.5 & 67.5 & 56.7 & 34.3 & 32.8 & 69.9 & 75.0 & 45.7 & 17.1 & 68.1 & 30.5 & 40.6 & 67.2 & 82.9 & 28.8 & 43.7 & 71.9 & \bf{62.0} & 62.8 & 58.2 & 54.2\\
   WSDDN-VGG$\_$M \cite{Ref:Bilen2016} & 65.1 & 63.4 & \bf{59.7} & 45.9 & 38.5 & 69.4 & 77.0 & 50.7 & 30.1 & 68.8 & 34.0 & 37.3 & 61.0 & 82.9 & 25.1 & 42.9 & \underline{\bf{79.2}} & 59.4 & 68.2 & 64.1 & 56.1\\
   WSDDN-VGG16 \cite{Ref:Bilen2016} & 65.1 & 58.8 & 58.5 & 33.1 & \bf{39.8} & 68.3 & 60.2 & \underline{\bf{59.6}} & 34.8 & 64.5 & 30.5 & 43.0 & 56.8 & 82.4 & 25.5 & 41.6 & 61.5 & 55.9 & 65.9 & 63.7 & 53.5\\
   WSDDN+context \cite{Ref:Kantorov2016} & \bf{83.3} & 68.6 & 54.7 & 23.4 & 18.3 & 73.6 & 74.1 & 54.1 & 8.6 & 65.1 & \bf{47.1} & \underline{\bf{59.5}} & 67.0 & 83.5 & \underline{\bf{35.3}} & 39.9 & 67.0 & 49.7 & 63.5 & 65.2 & 55.1\\
   \hline
   \methodname-VGG$\_$M & 81.7 & 72.9 & 56.5 & 31.4 & 36.3 & 75.6 & 81.6 & 57.0 & 7.3 & 74.7 & \bf{47.1} & 46.0 & \bf{78.2} & \bf{88.8} & 12.2 & 46.2 & 66.0 & 56.7 & 65.8 & 64.9 & 57.3\\
   \methodname-VGG16 & 81.7 & \bf{80.4} & 48.7 & \underline{\bf{49.5}} & 32.8 & \bf{81.7} & \bf{85.4} & 40.1 & \underline{\bf{40.6}} & \bf{79.5} & 35.7 & 33.7 & 60.5 & \bf{88.8} & 21.8 & \bf{57.9} & 76.3 & 59.9 & \bf{75.3} & \underline{\bf{81.4}} & \bf{60.6}\\
   \hline\hline
   OM+MIL+FRCNN \cite{Ref:Li2016} & 78.2 & 67.1 & 61.8 & 38.1 & 36.1 & 61.8 & 78.8 & \bf{55.2} & 28.5 & 68.8 & 18.5 & \bf{49.2} & 64.1 & 73.5 & 21.4 & 47.4 & 64.6 & 22.3 & 60.9 & 52.3 & 52.4\\
   WSDDN-Ens. \cite{Ref:Bilen2016} & 68.9 & 68.7 & \underline{\bf{65.2}} & 42.5 & 40.6 & 72.6 & 75.2 & 53.7 & \bf{29.7} & 68.1 & 33.5 & 45.6 & 65.9 & 86.1 & \bf{27.5} & 44.9 & \bf{76.0} & 62.4 & 66.3 & 66.8 & 58.0\\
   \hline
   \methodname-Ens. & 85.4 & 78.0 & 61.6 & 40.4 & 38.2 & 82.2 & 84.2 & 46.5 & 15.2 & 80.1 & 45.2 & 41.9 & 73.8 & 89.6 & 18.9 & 56.0 & 74.2 & 62.1 & 73.0 & 77.4 & 61.2\\
   \methodname-Ens.+FRCNN & \underline{\bf{85.8}} & \underline{\bf{82.7}} & 62.8 & \bf{45.2} & \underline{\bf{43.5}} & \underline{\bf{84.8}} & \underline{\bf{87.0}} & 46.8 & 15.7 & \underline{\bf{82.2}} & \underline{\bf{51.0}} & 45.6 & \underline{\bf{83.7}} & \underline{\bf{91.2}} & 22.2 & \underline{\bf{59.7}} & 75.3 & \underline{\bf{65.1}} & \underline{\bf{76.8}} & \bf{78.1} & \underline{\bf{64.3}}\\
\hline
\end{tabular}
}
\end{center}
\caption{CorLoc (in $\%$) for different methods on VOC 2007 trainval set.
The upper part shows results using a single model.
The lower part shows results of combing multiple models.}
\label{table:voc_2007_corloc}
\end{table*}

\subsubsection{The influence of instance classifier refinement}
\label{sec:influence_lp}

As in the blue line of Fig.~\ref{fig:ablation_refinement_time}, we can observe that compared with the basic network, even just refining instance classifier one time can boost the performance a lot (mAP from $29.5$ to $35.6$ and CorLoc from $49.9$ to $56.0$), which confirms the necessity of refinement.
If we refine the classifier multiple times, the results can be improved further.
But when refinement is implemented too many times, the performance tends to be saturated (the improvement from 2 times to 3 times is small).
Maybe this is because the network tends to converge so that the supervision of $3^{\textup{rd}}$ time is similar to $2^{\textup{nd}}$ time.
In the rest of this paper we only refine the classifier $3$ times.
Notice that in Fig.~\ref{fig:ablation_refinement_time}, the ``0 time'' is similar to the WSDDN \cite{Ref:Bilen2016} using SS as proposals.
Our result is a little worse than theirs ($30.9$ mAP in their paper), due to the different implementing platform and details.

\subsubsection{The influence of OICR}
\label{sec:influence_oil}

Fig.~\ref{fig:ablation_refinement_time} compares the results of different refinement times and different training strategies for classifier refinement.
As we can see, whether for our \methodname\ algorithm or the alternative strategy, results can be improved by refinement.
More importantly, compared with the alternatively refinement strategy, our \methodname\ can boost the performance consistently and significantly, which confirms the necessity of sharing representations.
Meanwhile, our method can also reduce the training time a lot, as it only requires to train a single model instead of training $K+1$ models for $K$ times refinement in the alternative strategy.

\subsubsection{The influence of weighted loss}
\label{sec:influence_wl}

We also study the influence of our weighted loss in Eq.~(\ref{equ:loss_refine_weighted}).
So here we train a network based on the Eq.~(\ref{equ:loss_refine}).
From Table~\ref{table:weight}, we can see that using the unweighted loss, the improvement from refinement is very scant, and the performance is even worse than the alternative strategy.
Using the weighted loss can achieve much better performance, which confirms our theory in Section~\ref{sec:oicr}.

\subsubsection{The influence of IoU threshold}
\label{sec:influence_it}

In previous experiments, we set the IoU threshold $I_{t}$ in the $12^{\textup{th}}$ line of Algorithm~\ref{alg:oicr} to $0.5$.
Here we conduct experiments to analyse the influence of $I_{t}$.
As in Fig.~\ref{fig:ablation_threshold}, $I_{t}=0.5$ outperforms other choices, and the results are not very sensitive to the $I_{t}$: when changing $I_{t}$ from $0.5$ to $0.6$, the performance only drops a little (mAP from $37.9$ to $37.8$, CorLoc maintains $57.3$).
Here we set $I_{t}$ to $0.5$ in other experiments.

\begin{table}[t]
\begin{center}
\footnotesize
\begin{tabular}{|l|c|c|}
   \hline
   Method & mAP ($\%$) & CorLoc ($\%$) \\
   \hline\hline
   WSDDN+context \cite{Ref:Kantorov2016} & 34.9 & 56.1 \\
   \hline
   \methodname-VGG$\_$M & 34.6 & 60.7\\
   \methodname-VGG16 & 37.9 & 62.1\\
   \hline
   \methodname-Ens. & 38.2 & 63.5 \\
   \methodname-Ens.+FRCNN & \bf{42.5} & \bf{65.6} \\
   \hline
\end{tabular}
\end{center}
\caption{Results for different methods on VOC 2012. Detailed per-class results can be found in Table~\ref{table:suppl_voc_2012_map} and Table~\ref{table:suppl_voc_2012_corloc} of the Supplementary Material.}
\label{table:voc_2012}
\end{table}

\begin{figure*}[!t]
\begin{center}
   \includegraphics[width=0.88\linewidth]{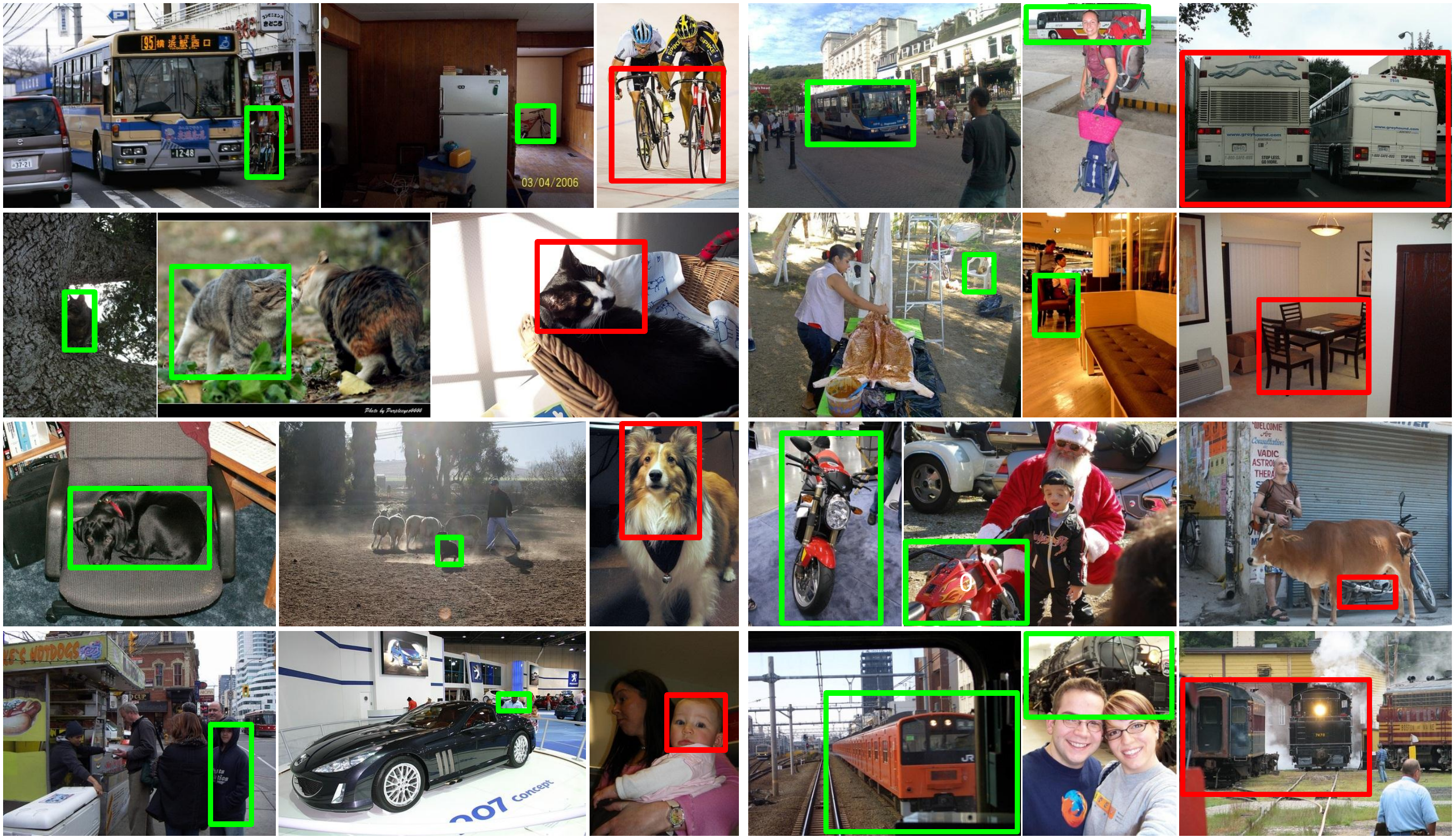}
\end{center}
   \caption{Some detection results for class bicycle, bus, cat, chair, dog, motorbike, person, and train.
   Green rectangle indicates success cases (IoU$>$0.5), and red rectangle indicates failure cases (IoU$<$0.5).}
\label{fig:visualize1}
\end{figure*}

\subsection{Comparison with other methods}
\label{sec:compar}

We report our results for each class on VOC 2007 and 2012 in Table~\ref{table:voc_2007_map}, Table~\ref{table:voc_2007_corloc}, and Table~\ref{table:voc_2012}.
Compared with other methods, our method achieves the state-of-the-art performance using single model, and even outperforms the results by combining multiple different models \cite{Ref:Bilen2016,Ref:Li2016}.
Specially, our methods achieves much better performance than the method by Bilen and Vedaldi \cite{Ref:Bilen2016} using the same CNN model.
Notice that \cite{Ref:Bilen2016} not only uses the weighted pooling as we stated in Section~\ref{sec:midn}, but also combines the objectness measure of EdgeBoxes \cite{Ref:Zitnick2014} and the spatial regulariser, which is much complicated than our basic MIDN.
We believe that our performance can be improved by choosing better basic detection network, like the complete network in \cite{Ref:Bilen2016} and using the context information \cite{Ref:Kantorov2016}.
As reimplementing their method completely is trivial, here we only choose the simplest architecture in \cite{Ref:Bilen2016}.
Even in this simplified case, our method can achieve very promising results.
We also show some visualization comparisons among the WSDDN \cite{Ref:Bilen2016}, the WSDDN+context \cite{Ref:Kantorov2016}, and our method in Fig.~\ref{fig:suppl_visualize_comparison} of the Supplementary Material.

Our results can also be improved by combing multiple models.
As shown in the tables, if we simply sum up the scores produced by the VGG$\_$M model and VGG16 model (\methodname-Ens. in tables), there is little improvement.
Also, as mentioned in Section~\ref{sec:exp_setup}, we train a FRCNN detector using top-scoring proposals produced by \methodname-Ens. as ground truths (\methodname-Ens.+FRCNN in tables).
As we can see, the performance can be improved further.

Though our method significantly outperforms other methods for some class, like ``bicyle'', ``bus'', ``motorbike'', etc, the performance is poor for classes like ``cat'', ``dog'', and ``person''.
For analysis, we visualize some success and failure detection results on VOC 2007 trainval by \methodname-Ens., as in Fig.~\ref{fig:visualize1}.
We can observe that, our method is robust to the size and aspect of objects, especially for rigid objects.
The main failures for these rigid objects are always due to overlarge boxes that not only contain objects, but also include their adjacent similar objects.
For non-rigid objects like ``cat'', ``dog'', and ``person'', they are always with great deformation, while there is less deformation of their most representative parts (like head), so our detector is still inclined to find these parts.
An ideal solution is yet wanted because there is still room for improvement.

\section{Conclusion}
\label{sec:con}

In this paper, we present a novel algorithm framework for weakly supervised object detection.
Different from traditional approaches in this field, our method integrates a basic multiple instance detection network and multi-stage instance classifiers into a single network.
Moreover, we propose an online instance classifier refinement algorithm to train the proposed network end-to-end.
Experiments show substantial and consistent improvements by our method.
Our learning algorithm is potential to be applied in many other weakly supervised visual learning tasks.
In the future, we will explore other cues such as instance visual similarity for performing instance classifier refinement better.

\paragraph{Acknowledgements}
This work was partly supported by NSFC (No. 61503145, No. 61572207, No. 61573160) and the CAST Young Talent Supporting Program.

{\small
\bibliographystyle{ieee}
\bibliography{egbib}
}

\clearpage
\renewcommand{\figurename}{Supplementary Figure}
\renewcommand{\tablename}{Supplementary Table}
\setcounter{section}{0}
\setcounter{figure}{0}
\setcounter{table}{0}

{\LARGE Supplementary Material}
\maketitle


Here are the supplementary materials for ``Multiple Instance Detection Network with Online Instance Classifier Refinement''.
We provide detailed per-class results on VOC 2012, and some visualization comparisons among the WSDDN \cite{Ref:Bilen2016}, the WSDDN+context \cite{Ref:Kantorov2016}, and our method (OICR).

\section{Per-class results on VOC 2012}
\label{sec:suppl_perclass_res}

\begin{table*}[t]
\begin{center}
\resizebox{\linewidth}{!}{
\begin{tabular}{|@{}L{3.6cm}|*{20}{x}|x|}
   \hline
   Method & aero & bike & bird & boat & bottle & bus & car & cat & chair & cow & table & dog & horse & mbike & person & plant & sheep & sofa & train & tv & mAP \\
   \hline\hline
   WSDDN+context \cite{Ref:Kantorov2016} & 64.0 & 54.9 & 36.4 & 8.1 & 12.6 & 53.1 & 40.5 & \bf{28.4} & 6.6 & 35.3 & \bf{34.4} & \bf{49.1} & 42.6 & 62.4 & \bf{19.8} & 15.2 & 27.0 & 33.1 & 33.0 & 50.0 & 35.3\\
   \hline
   \methodname-VGG$\_$M$^{\dag}$ & 64.4 & 50.6 & 34.8 & 16.7 & 16.5 & 49.7 & 44.8 & 20.4 & 5.0 & 39.0 & 18.2 & 46.2 & 50.3 & 64.3 & 3.4 & 15.1 & 32.4 & 38.5 & 36.3 & 45.1 & 34.6\\
   \methodname-VGG16$^{\ddag}$ & 67.7 & 61.2 & 41.5 & 25.6 & 22.2 & 54.6 & 49.7 & 25.4 & \bf{19.9} & 47.0 & 18.1 & 26.0 & 38.9 & 67.7 & 2.0 & 22.6 & 41.1 & 34.3 & 37.9 & 55.3 & 37.9\\
   \hline
   \methodname-Ens.$^{\S}$ & 68.4 & 58.6 & 39.9 & 23.4 & 21.3 & 52.8 & 48.7 & 23.5 & 13.5 & 44.8 & 22.0 & 36.5 & 47.6 & 68.3 & 2.6 & 21.7 & 39.1 & \bf{39.6} & \bf{38.2} & 52.7 & 38.2\\
   \methodname-Ens.+FRCNN$^{\natural}$ & \bf{71.4} & \bf{69.4} & \bf{55.1} & \bf{29.8} & \bf{28.1} & \bf{55.0} & \bf{57.9} & 24.4 & 17.2 & \bf{59.1} & 21.8 & 26.6 & \bf{57.8} & \bf{71.3} & 1.0 & \bf{23.1} & \bf{52.7} & 37.5 & 33.5 & \bf{56.6} & \bf{42.5}\\
\hline
\end{tabular}
}
\end{center}
\caption{Average precision (in $\%$) for different methods on VOC 2012 test set.
$^{\dag}$\url{http://host.robots.ox.ac.uk:8080/anonymous/PGSNG5.html}
$^{\ddag}$\url{http://host.robots.ox.ac.uk:8080/anonymous/6ASJ4I.html}
$^{\S}$\url{http://host.robots.ox.ac.uk:8080/anonymous/UEZKOR.html}
$^{\natural}$\url{http://host.robots.ox.ac.uk:8080/anonymous/XP6BJ7.html}
}
\label{table:suppl_voc_2012_map}
\end{table*}

The detailed per-class results on VOC 2012 can be viewed in Supplementary Table~\ref{table:suppl_voc_2012_map} and Supplementary Table~\ref{table:suppl_voc_2012_corloc}.
Obviously our method outperforms the previous state-of-the-art \cite{Ref:Kantorov2016} by a large margin.
Similar to results on VOC 2007, our results are better than \cite{Ref:Kantorov2016} for rigid objects such as ``bicycle'', ``car'', and ``motorbike'', but worse than \cite{Ref:Kantorov2016} for non-rigid objects ``cat'', ``dog'', and ``person''.
This is because non-rigid objects are always with great deformation.
Our method tends to detect the most discriminative parts of these objects (like head).
The \cite{Ref:Kantorov2016} considers more context information thus can deal with these classes better.

\begin{table*}[t]
\begin{center}
\resizebox{\linewidth}{!}{
\begin{tabular}{|@{}L{3.6cm}|*{20}{x}|x|}
   \hline
   Method & aero & bike & bird & boat & bottle & bus & car & cat & chair & cow & table & dog & horse & mbike & person & plant & sheep & sofa & train & tv & mean \\
   \hline\hline
   WSDDN+context \cite{Ref:Kantorov2016} & 78.3 & 70.8 & 52.5 & 34.7 & 36.6 & 80.0 & 58.7 & \bf{38.6} & 27.7 & 71.2 & 32.3 & 48.7 & 76.2 & 77.4 & \bf{16.0} & 48.4 & 69.9 & 47.5 & \bf{66.9} & 62.9 & 54.8\\
   \hline
   \methodname-VGG$\_$M & 84.8 & 79.7 & 66.0 & 44.4 & 40.2 & 75.6 & 72.2 & 32.3 & 20.0 & 82.5 & 46.2 & \bf{64.1} & \bf{84.3} & 87.5 & 12.3 & 48.6 & 75.8 & \bf{63.9} & 62.5 & 70.3 & 60.7\\
   \methodname-VGG16 & 86.2 & 84.2 & 68.7 & 55.4 & 46.5 & 82.8 & 74.9 & 32.2 & 46.7 & 82.8 & 42.9 & 41.0 & 68.1 & 89.6 & 9.2 & 53.9 & 81.0 & 52.9 & 59.5 & 83.2 & 62.1\\
   \hline
   \methodname-Ens. & 86.7 & 85.1 & 69.5 & 57.2 & 47.4 & 81.2 & 76.2 & 32.0 & 34.0 & 84.8 & \bf{47.9} & 50.4 & 82.0 & 88.6 & 10.4 & 55.5 & 77.9 & 62.6 & 60.2 & 80.8 & 63.5\\
   \methodname-Ens.+FRCNN & \bf{89.3} & \bf{86.3} & \bf{75.2} & \bf{57.9} & \bf{53.5} & \bf{84.0} & \bf{79.5} & 35.2 & \bf{47.2} & \bf{87.4} & 43.4 & 43.8 & 77.0 & \bf{91.0} & 10.4 & \bf{60.7} & \bf{86.8} & 55.7 & 62.0 & \bf{84.7} & \bf{65.6}\\
\hline
\end{tabular}
}
\end{center}
\caption{CorLoc (in $\%$) for different methods on VOC 2012 trainval set.}
\label{table:suppl_voc_2012_corloc}
\end{table*}

\section{Visualization comparisons}

We show some visualization comparisons among the WSDDN \cite{Ref:Bilen2016}, the WSDDN+context \cite{Ref:Kantorov2016}, and our method in Supplementary Fig.~\ref{fig:suppl_visualize_comparison}.
From this visualization and results from tables, we can observe that for classes such as aeroplane, bike, car, \etc, our method tends to provide more accurate detections, whereas other two methods sometimes fails to produce boxes that are overlarge or only contain parts of objects (the first four rows in Supplementary Fig.~\ref{fig:suppl_visualize_comparison}).
But for some classes such as person, our method always fails to detect only parts of objects (the fifth row in Supplementary Fig.~\ref{fig:suppl_visualize_comparison}).
Because considering context information sometimes help the detection (as in WSDDN+context \cite{Ref:Kantorov2016}), we believe our method can be further improved by incorporating context information into our framework.

All these three methods (actually almost all weakly supervised object detection methods) suffers from two problems: producing boxes that not only contain the target object but also include their adjacent similar objects, or only detecting parts of object for objects with deformation (the last row in Supplementary Fig.~\ref{fig:suppl_visualize_comparison}).
An ideal solution for these problems is yet wanted.

\begin{figure*}[!t]
\begin{center}
   \includegraphics[width=\linewidth]{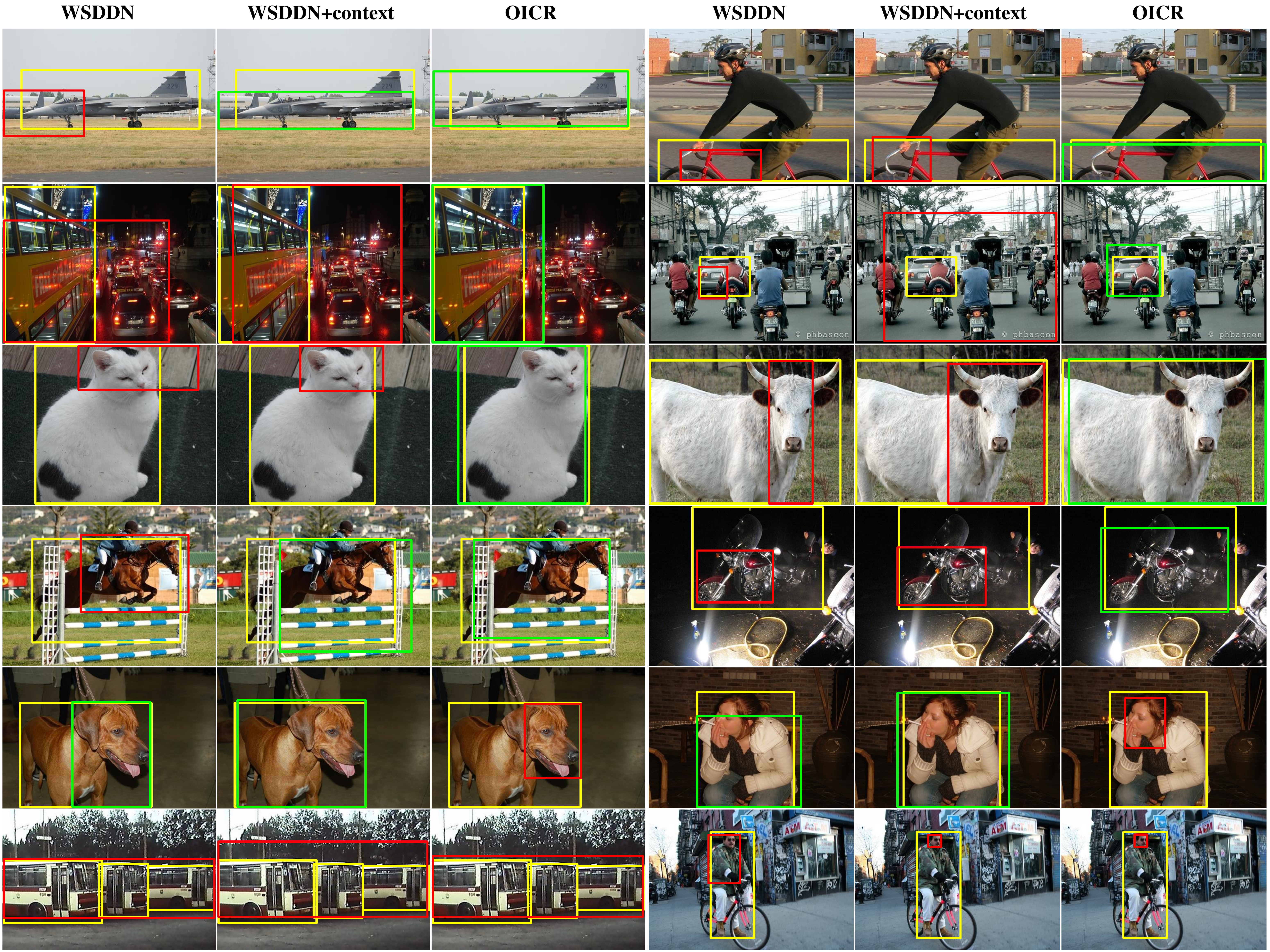}
\end{center}
   \caption{Some visualization comparisons among the WSDDN \cite{Ref:Bilen2016}, the WSDDN+context \cite{Ref:Kantorov2016}, and our method (OICR).
   Green rectangle indicates success cases (IoU$>$0.5), red rectangle indicates failure cases (IoU$<$0.5), and yellow rectangle indicates ground truths.
   The first four rows show examples that our method outperforms other two methods (with larger IoU).
   The fifth row shows examples that our method is worse than other two methods (with smaller IoU).
   The last row shows failure examples for both three methods.
   }
\label{fig:suppl_visualize_comparison}
\end{figure*}

\end{document}